\documentclass[10pt,twocolumn,letterpaper]{article}

\usepackage{cvpr}              % To produce the CAMERA-READY version
\usepackage{times}
\usepackage{epsfig}
\usepackage{graphicx}
\usepackage{amsmath}
\usepackage{amssymb}
\usepackage{mathtools}
\usepackage{multirow}
\usepackage{enumitem}
\usepackage[accsupp]{axessibility} % Improves PDF readability for those with disabilities.
\usepackage{color}

\def\orange#1{\textcolor[rgb]{0.84,0.61,0.0}{#1}}
\def\violet#1{\textcolor[rgb]{0.59,0.45,0.65}{#1}}

\newcommand{\keypoint}[1]{\vspace{0.08cm}\noindent\textbf{#1}\quad}
\newcommand{\cut}[1]{}
\usepackage{dsfont}
\usepackage{algorithm}
\usepackage{algpseudocode}
\usepackage{makecell}
\usepackage{lipsum}
\usepackage{amsmath}
\usepackage{titling}
\usepackage{wrapfig}
\usepackage{bm}
\usepackage{pifont}% http://ctan.org/pkg/pifont
\newcommand{\xmark}{\ding{55}}%

\usepackage{multirow}

\algnewcommand\algorithmicforeach{\textbf{for each}}
\algdef{S}[FOR]{ForEach}[1]{\algorithmicforeach\ #1\ \algorithmicdo}

\makeatother
\usepackage[pagebackref=true,breaklinks=true,colorlinks,bookmarks=false]{hyperref}

\def\cvprPaperID{6038} % *** Enter the CVPR Paper ID here

\begin{document}

%%%%%%%%% TITLE
\title{Sketch3T: Test-Time Training for Zero-Shot SBIR}

\author{
Aneeshan Sain\textsuperscript{1,2}  \hspace{.2cm} 
Ayan Kumar Bhunia\textsuperscript{1} \hspace{.3cm}  
Vaishnav Potlapalli\thanks{Interned with SketchX}  \hspace{.3cm}  
Pinaki Nath Chowdhury\textsuperscript{1,2}   \\  
Tao Xiang\textsuperscript{1,2}\hspace{.2cm}  
Yi-Zhe Song\textsuperscript{1,2} \\
\textsuperscript{1}SketchX, CVSSP, University of Surrey, United Kingdom.  \\
\textsuperscript{2}iFlyTek-Surrey Joint Research Centre on Artificial Intelligence.\\
{\tt\small \{a.sain, a.bhunia, p.chowdhury,  t.xiang, y.song\}@surrey.ac.uk
% ; pvaishnav2718@gmail.com
} 
}
\date{}
\maketitle
% \ifcvprfinal\thispagestyle{empty}\fi

%%%%%%%%% ABSTRACT
\begin{abstract}
\vspace{-0.2cm}
{
Zero-shot sketch-based image retrieval typically asks for a trained model to be applied as is to unseen categories. In this paper, we question to argue that this setup by definition is not compatible with the inherent abstract and subjective nature of sketches -- the model might transfer well to new categories, but will not understand sketches existing in different test-time distribution as a result. We thus extend ZS-SBIR asking it to transfer to both categories and sketch distributions. Our key contribution is a test-time training paradigm that can adapt using just one sketch. Since there is no paired photo, we make use of a sketch raster-vector reconstruction module as a self-supervised auxiliary task. To maintain the fidelity of the trained cross-modal joint embedding during test-time update, we design a novel meta-learning based training paradigm to learn a separation between model updates incurred by this auxiliary task from those off the primary objective of discriminative learning. Extensive experiments show our model to outperform state-of-the-arts, thanks to the proposed test-time adaption that not only transfers to new categories but also accommodates to new sketching styles. 
}
\end{abstract}

%%%%%%%%% BODY TEXT

%%%%%%%%% BODY TEXT
\vspace{-0.2cm}
\section{Introduction}\label{sec:intro}

Sketch-based image retrieval (SBIR) is by now a well-established topic in the vision community \cite{dutta2019semantically, dey2019doodle}. Research efforts have mainly focused on addressing the sketch-photo domain gap, incurred by abstraction \cite{umar2019goal}, drawing style \cite{sain2021stylemeup} and stroke saliency \cite{ha2017neural}. Despite great strides made, the field remains plagued by the data scarcity problem -- sketches are notoriously difficult to collect~\cite{bhunia2021more,bhunia2022diy}. 

\begin{figure}[!t]
\begin{center}
  \includegraphics[width=1\linewidth]{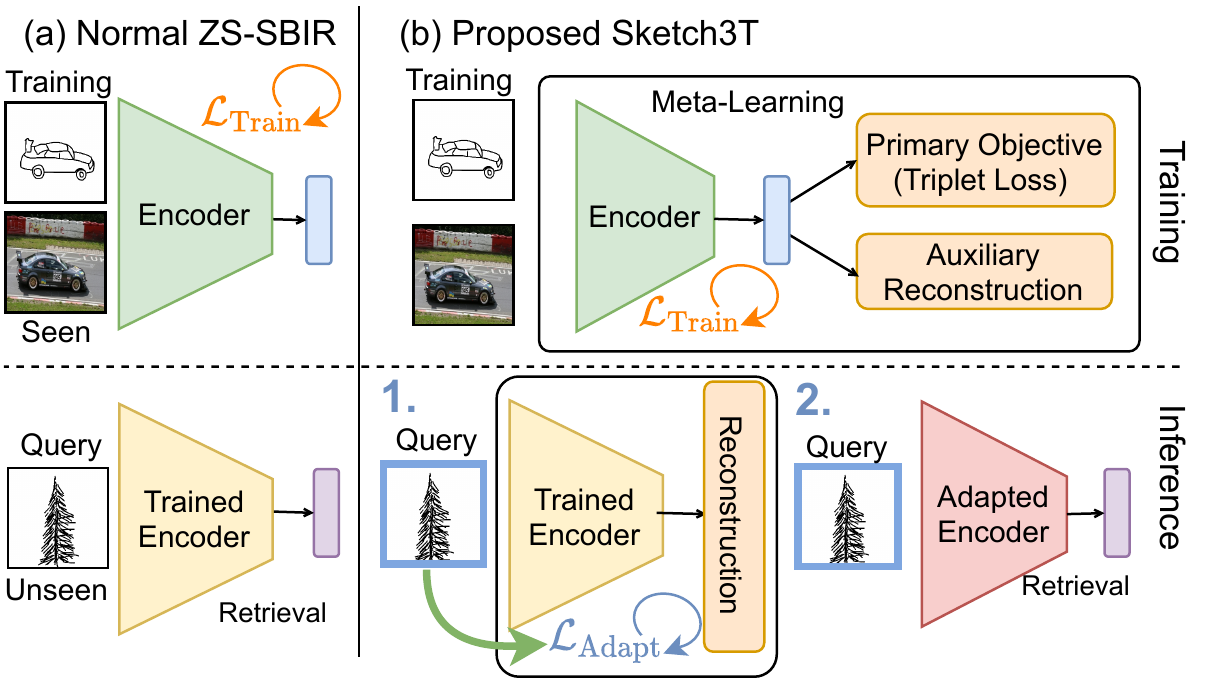}
\end{center}
    \vspace{-0.1cm}
  \caption{
  {Normal ZS-SBIR methods obtain lower accuracies as they retrieve from unseen data using model weights trained on seen data. During inference, our model (Sketch3T) adapts to the test distribution via an auxiliary task, before retrieval, scoring better.}
  }
\label{fig:opener}
\vspace{-0.1cm}
\end{figure}

Zero-shot SBIR (ZS-SBIR) in particular represents the main body of work behind this push for addressing data scarcity. It specifically examines the scarcity issue from a category transfer perspective, and strives for utilising sketch-photo pairs from seen categories to train a model that could be \textit{directly} applied on those unseen (see Fig. \ref{fig:opener}(a)). 

In this paper, we question this otherwise commonly accepted setup at definition level. We importantly argue that the very assumption of being able to apply a trained model \textit{as is} to unseen categories, is by definition \textit{incompatible} with the inherent subjective nature of sketch data. This largely results in a model that might well understand the semantic category shift, but not acute to changes in sketching style and abstraction level (both being prevalent problems in sketch \cite{sain2021stylemeup}). Alleviating this problem is particularly crucial for the practical adaption of SBIR, as otherwise retrieval performance will incur a significant drop -- a system that understands ``my" sketches, might not understand ``yours".

This paper thus extends the conventional definition of ZS-SBIR to embrace this new problem, i.e., a new ZS-SBIR framework that (i) not only transfers knowledge on to unknown categories, (ii) but also adapts to the unique style of new sketches. 
We implement this by adopting a test-time-training framework that adapts to new categories and new styles \textit{at inference time}. That is, instead of anticipating the distribution shifts via normal training, we intend to learn them at test time.
The beauty of our solution lies in that we achieve a higher accuracy without any additional training sketch-photo pairs, but with just a \textit{single query sketch}, no more than what is required in a typical ZS-SBIR setup (Fig. \ref{fig:opener}). It follows that this single sketch will first adjust the model to unseen style and category, and then use again as query to retrieve using the updated model, \textit{all test time}.

Implementing this test-time-training framework despite intuitive is not trivial. There are two major challenges: Firstly, we have access to only the query sketches \textit{during inference}, without any label or paired photo for supervision. Secondly, this test-time update should not degrade the joint embedding (that conducts retrieval) which has been learned using sketch-pairs. Solution to the first issue requires a task where labels can be obtained freely/synthetically during inference itself. Here we make clever use of the vectorised nature of sketches, and utilise a self-supervised task of sketch-raster to sketch-vector translation \cite{sketch2vec} {to update the feature-extractor at inference}. It follows that via this translation operation, the model adapts itself to the new style/category of the test sketch.

The second issue gets tackled at model design. In particular, we consolidate the said sketch self-reconstruction module as an auxiliary task within a meta-learning framework \cite{li2017meta}. It follows that the model is meta-learned in a way, such that updates on the auxiliary task only happens in the inner loop, which then prevents it from distorting the joint embedding space whose updates occurs elsewhere in the outer loop via a triplet loss.
This training strategy essentially ensures that the trained model now \textit{knows} how to accommodate the auxiliary task loss without affecting the latent space too adversely, and accordingly defends itself against test-time updates from the sketch self-reconstruction auxiliary task.

\begin{figure*}[t]
\begin{center}
\centering
\includegraphics[width=1\linewidth]{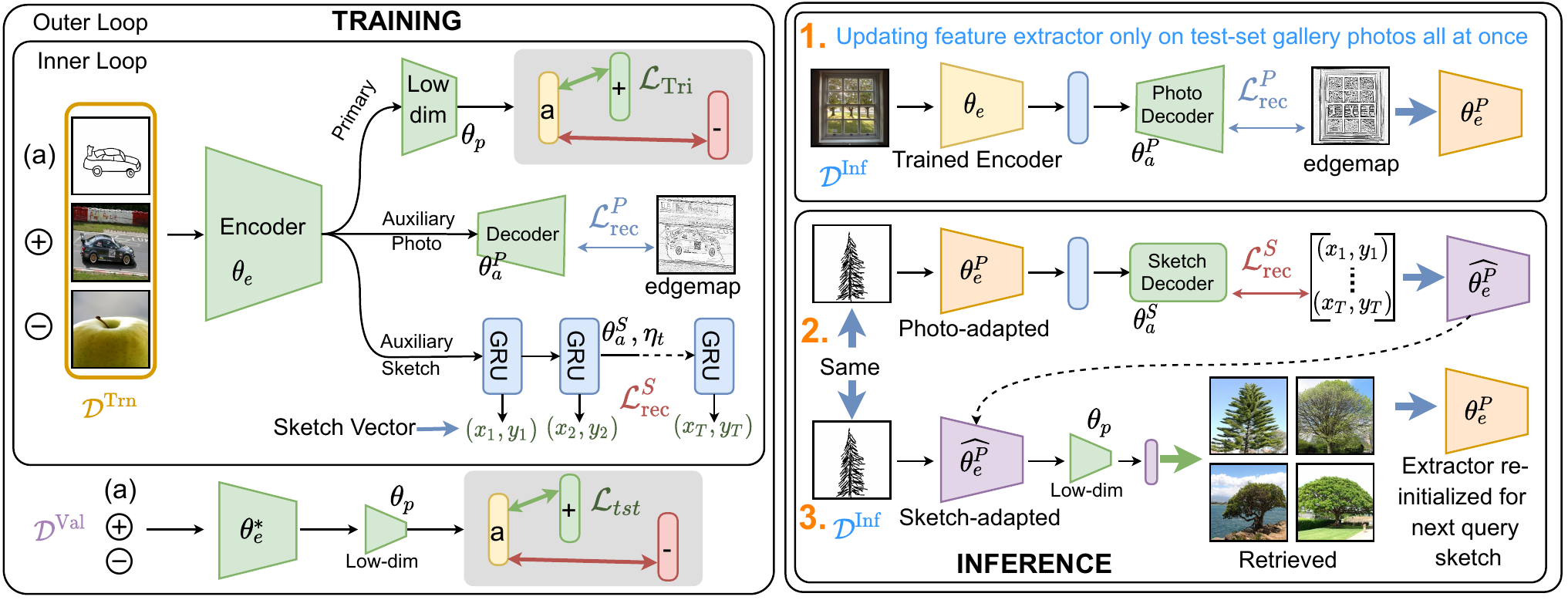} % 344 vs 338
\end{center}
\vspace{-.6cm}
  \caption{Our Framework. Our model is trained (left) on primary and auxiliary tasks, meta-learning stroke-weights. During inference (right) the model first updates (optionally) on the test-set photo distribution, followed by sketch-specific test-time training for retrieval.}
  \label{fig:framework}
\vspace{-.5cm}
\end{figure*}

More specifically, our framework shares a feature extractor amongst three diverging branches (Fig. \ref{fig:framework} (left)): (i) a primary branch learns the cross-modal embedding over a triplet loss \cite{yu2016sketch} using paired sketch-photo information, (ii) an auxiliary sketch branch that focuses on self-modal reconstruction to update and condition the shared feature-extractor towards better sketch-encoding, and (iii) an auxiliary photo branch, where we use photo-to-edgemap translation to condition the photo features. Having this photo branch also presents the \textit{option} of updating the model on the \textit{unseen} test-set photo-gallery to yield better photo features for retrieval, however is not compulsory. Note that only the auxiliary sketch branch get updated (and hence the shared feature extractor) at test-time upon a query sketch. 

Our contributions are:
(a) We offer a fresh extension on the ZS-SBIR paradigm, by proposing a novel test-time training framework that dynamically adapts a trained encoder to new sketches
(b) To retrain transferable cross-modal embedding knowledge during inference, we propose a meta-learning framework that integrates primary discriminative learning with auxiliary tasks, such that updates from the latter are constrained towards benefiting the primary objective. 
(c) Extensive experiments and ablation confirm our method to be superior to existing state-of-the-arts.

\vspace{-0.1cm}
\section{Related Works}
\vspace{-0.15cm}

\noindent \textbf{Sketch Based Image Retrieval (SBIR):}
SBIR involves finding an image corresponding to a given query-sketch. Aiming to retrieve photos of the same category, category-level SBIR \cite{sangkloy2016sketchy, collomosse2017sketching} began with using handcrafted descriptors~\cite{tolias2017asymmetric} like SIFT \cite{lowe1999object}, Gradient Field HOG \cite{hu2013performance}, Histogram of Edge Local Orientations \cite{saavedra2014sketch} or Learned Key Shapes \cite{saavedra2015sketch}, for constructing local \cite{hu2013performance} or global \cite{qi2015making} joint photo-sketch representations. Shifting to deep-learning, methods \cite{liu2017deep, yang2020s, collomosse2019livesketch} usually trained Siamese-like networks to fetch similar photos over a distance-metric in a cross-modal joint embedding space, over ranking losses~\cite{collomosse2017sketching}. Contemporary research include embedding sketch features to binary hash-codes \cite{liu2017deep, zhang2018generative} for computational ease. Sketch as a query \cite{pinaki2022PartialSBIR} however, prides in its ability to model fine-grained details. Research thus advanced to \textit{\textbf{f}ine-\textbf{g}rained} SBIR \cite{song2018learning, pang2019generalising, bhunia2022subset} beginning with deformable-part models~\cite{li2014fine}. Aided by new datasets \cite{song2017deep, yu2016sketch}, FG-SBIR flourished with the introduction of triplet-ranking models~\cite{yu2016sketch}, learning a joint sketch-photo manifold. Attention mechanisms along with higher-order losses \cite{song2017deep}, hybrid generative-discriminative cross-domain image generation \cite{pang2017cross}, textual tags \cite{song2017fine} and mixed-modal jigsaw solving based pre-training strategy \cite{pang2020solving}, enhanced it further. While Sain \etal~\cite{sain2020cross} discovered cross-modal hierarchy in sketches, Bhunia \etal~\cite{bhunia2020sketch} employed reinforcement learning in an early retrieval scenario. Although further works have addressed low-resource data via semi-supervised learning \cite{bhunia2021more}, or style-diversity in sketches via meta-learning disentanglement \cite{sain2021stylemeup}, training during inference to bridge the train-test data distribution gap, remains unseen in SBIR.

\noindent \textbf{Zero-Shot Learning:} %Zero-shot SBIR:
To deal with the data scarcity, a separate branch of literature has evolved within the SBIR pipeline that aims to generalise the knowledge learned from \textit{seen} training classes to \textit{unseen} testing categories. A zero-shot (ZS) SBIR pipeline was first introduced by Yelamarthi \etal \cite{yelamarthi2018zero}, with an aim to minimise sketch-photo domain gap by approximating photo features from given sketches via image-to-image translation, thus aligning sketch-photo features jointly to generalise onto unseen classes. In contrast later works ~\cite{dey2019doodle, dutta2019semantically} used semantic representation (word2vec) of class labels to learn a joint manifold capable of semantic transfer to unseen categories. While, \cite{dutta2019semantically} used adversarial training to align sketch, photo and semantic representation, \cite{dey2019doodle} employed a gradient reversal layer to minimise sketch-photo domain gap. Other works include preserving training knowledge via knowledge distillation \cite{liu2019semantic} to improve generalisability, and alleviating sketch-image heterogeneity via Kronecker fusion layer with graph convolution \cite{shen2018zero}, thus enhancing semantic relations among data towards a generative hashing scheme for ZS-SBIR.

While earlier ZS-SBIR methods fixed model weights after training on seen classes, we advocate for one that adapts to novel classes during inference. Please note that this `adaptation protocol' must \textit{not} be confused with that of few-shot learning \cite{snell2017prototypical, wang2020generalizing} which considers access to a few labelled samples. We however have \textit{no access} to labelled data from unseen categories under ZS-SBIR setup. To adapt to unseen classes, we thus employ a self-supervised task for sketch and photo branch each, whose loss could be computed using labels that can be obtained freely/synthetically. Additionally, this self-supervised objective should imbibe knowledge of unseen classes via a few gradient update steps within a reasonable remit of edge device deployment.

\noindent \textbf{Self-supervised Auxiliary Tasks:}
{
Constrained by the absence of labels during inference, our choice of task for test-time training should be a self-supervised one. Self-supervision involves designing pretext tasks that can learn semantic information without human annotations \cite{jing2020selfSup}, such as image colorization \cite{zhang2016colorful},  super-resolution \cite{ledig2017photo}, frame order recognition \cite{misra2016shuffle}, solving jigsaw puzzles \cite{noroozi2016unsupervised,pang2020solving}, image in-painting \cite{pathak2016context}, relative patch location prediction \cite{doersch2015unsupervised}, etc. Importantly, Asano et al. \cite{asano2019surprising} shows self-supervised learning on a single image, can produce low level features that generalise well. However, they use complex tuple selection \cite{misra2016shuffle} or patch-sampling strategies \cite{noroozi2016unsupervised,pang2020solving} and relation-operations which leads to complex design issues in batch size, sampling strategies, or data-balancing that need tuning. We thus opt for simple self-modal reconstruction for test-time training. As an auxiliary task during training, it should improve robustness of the primary task~\cite{hendrycks2019using}, like rotation prediction~\cite{sun2020test} or via entropy minimisation~\cite{wang2021tent}. Similar notion has been used in few-shot learning \cite{su2019boosting}, domain generalization \cite{carlucci2019domain}, and unsupervised domain adaptation \cite{sun2019unsupervised,liang2020we}. Following suit, we use raster-to-vector decoding and image-to-edgemap translation as auxiliary tasks for sketch and photo branch respectively during training.
}

\noindent \textbf{Meta Learning:}
{
This aims to extract transferable knowledge from a series of related tasks, to help adapt to unseen tasks with a few training samples~\cite{finn2017MAML,vinyals2016matching}. 
% One approach would be to learn a good initialization, that would help in adapting to new tasks quickly \cite{snell2017prototypical,}.
Broadly speaking these algorithms fall in three groups. \textit{Metric}-based methods \cite{snell2017prototypical,sung2018learning} strive to create a metric space where learning is efficient with just a few samples. \textit{Memory network} based approaches \cite{oreshkin2018tadam} attain knowledge across tasks, to generalise on the unseen task. \textit{Optimization}-based techniques \cite{finn2017MAML, nichol2018first, sun2019meta} optimises a model, such that it can adapt to any test data quickly. Specifically, we use the popular model-agnostic meta-learning (MAML) algorithm \cite{finn2017MAML} (enhanced to MAML++~\cite{antoniou2018train}), due its compatibility with any model trained via gradient descent, and diverse application range with several variants \cite{nichol2018first, antoniou2018train, rusu2019LEO, stylemeup}. Besides using it to condition our model in a test-time scenario during training, we modify it to meta-train learnable stroke-specific weights for reconstruction, like learning rates in MetaSGD~\cite{li2017meta}.
}

\vspace{-0.2cm}
\section{Background Study}
\vspace{-0.15cm}
\keypoint{Baseline SBIR:} \label{baseSBIR}
Sketch Based Image Retrieval aims at retrieving an image pertaining to a sketch query. For categorical SBIR ~\cite{collomosse2017sketching}, the image is retrieved from a gallery having images of different classes, and ideally belongs to the \textit{same category} as that of the sketch.
Formally, our model learns an embedding function, $\mathcal{F}_\theta(\cdot): \mathbb{R}^{H \times W \times 3} \rightarrow \mathbb{R}^d$, mapping a rasterised sketch or photo $I$ to a d-dimensional feature. Given a gallery of $G = \{C_i\}_{i=1}^M$ categories, having $N_i$ photos each, our core SBIR model obtains a list of photo ($p$) features $\widehat{G} = \mathcal{F}_\theta(\{p_j^{C_i}\}_{j=1}^{N_i}|_{i=1}^{M})$. 
Thereafter, pairwise distances are calculated and corresponding images are retrieved over a precision metric~\cite{dey2019doodle}.
A state-of-the-art CNN $(\mathcal{F}_\theta(\cdot))$ extracts features of query sketch ($S$), matching photo ($P^+$) and an unmatched one ($P^-$) which are trained on a triplet loss objective \cite{yu2016sketch}, where minimising the loss signifies bringing the sketch-feature ($f_{S}$) closer to the positive photo-feature ($f_{P^+}$) while distancing it from the negative ($f_{P^-}$) one in the joint embedding space.
\vspace{-0.2cm}
\begin{equation}
    \begin{aligned}
    \mathcal{L}_{\text{Tri}}^\theta = \max \{0, m + \delta(f_S, f_{P^+}) - \delta(f_S, f_{P^-})\}
    \end{aligned}
    \label{equ:baseTrip}
\end{equation}
where, $\delta(a,b)$ = $||a-b||^2$, is a distance metric and $m$ is a margin hyperparameter, obtained empirically.

\keypoint{Test-time Training:}
{
During inference, given a query-sketch ($S^\mathcal{T}$), the trained feature extractor ($\theta_e$) is updated based on a proxy-task to adapt to this specific test-sample. This task must be self-supervised to be free of label-cost. 
Features then extracted by the updated model ($\widehat{\theta_e}$) are used to calculate pairwise distances for retrieval. Constrained by the unavailability of labels during inference, self-modal reconstruction is a common task-choice. More importantly, this task is used during training as well, as an auxiliary task to improve the model's primary objective. Consequently, we have three sets of parameters: the shared feature encoder ($\theta_e$), the exclusive primary-task parameters ($\theta_p$) and auxiliary-task parameters ($\theta_{a}$). During test-time training, the common feature extractor is updated using the auxiliary task loss ($\mathcal{L}_\text{aux}$) to perform primary task on $S^\mathcal{T}$ as,
\vspace{-0.2cm}
\begin{equation}
    \begin{aligned}
    \operatorname*{min}_{\theta_e} \; \mathcal{L}_\text{aux} (S^\mathcal{T}; \theta_e, \theta_a)
    \; , \;
    f_{S^\mathcal{T}} = \mathcal{F}_{\widehat{\theta_e}, \theta_p} (S^\mathcal{T})
    \end{aligned}
    \label{equ:mainTTT}
\vspace{-0.25cm}
\end{equation}
After operating on $S^\mathcal{T}$, $\widehat{\theta_e}$ is discarded as standard practice, and feature extractor is re-initialised with $\theta_e$ for a fresh adaptation on the next test sample. 
}

\vspace{-0.1cm}
\section{Methodology}
\vspace{-0.15cm}
\keypoint{Overview:}   
{
We aim to devise a SBIR framework that learns to alleviate test-train distribution gap by aligning a trained model to the test-data distribution, thus achieving better retrieval accuracy. To this end, we design a SBIR model which is trained in a meta-learning framework, augmented via auxiliary training and enhanced (for the first time) via a test-time training paradigm. First, a feature extractor (Sec. \ref{baseSBIR}(i)) encodes a query-sketch ($S$), its matched photo ($P^+$), and an unmatched one ($P^-$) to obtain features $f_{S}$, $f_{P^+}$ and $f_{P^-}$ all $\in \mathbb{R}^d$ respectively using $\mathcal{F}_{\theta_e}(\cdot)$. Thereafter the model is trained in two branches (Fig. \ref{fig:framework}). While the \textit{primary} branch ($\theta_p$) instils cross-modal discriminative knowledge via triplet loss objective on those three features (Eq. \ref{equ:baseTrip}), the \textit{auxiliary} branch ($\theta_a$) is trained on a self-modal reconstruction loss to improve primary task. Accordingly, we perform raster-to-vector decoding for sketch, and photo-to-edgemap translation for photo, to obtain reconstruction loss. Furthermore we associate learnable weights to every sketch-stroke which are meta-learned along with other modules in a meta-learning framework, to imbibe the knowledge of relative importance of strokes during reconstruction towards a better retrieval accuracy. For \textit{every} test sample during inference, the feature-extractor is first \textit{initialised} with trained parameters ($\theta_e$). Following Sec. \ref{baseSBIR}(ii), it is updated via reconstruction loss to adapt to the test distribution. Features extracted by the \textit{updated} model are used for retrieval.
}

\vspace{-0.2cm}
\subsection{Model Architecture}\label{basics_maml}
\vspace{-0.2cm}

Our pipeline starts with a feature-extractor ($\theta_e$), which bifurcates into a primary branch ($\theta_p$) focused at cross-modal discriminative learning, and an auxiliary branch ($\theta_a$) for self-reconstruction task. The feature extractor (shared between two branches) first encodes a photo or sketch-image into a d-dimensional feature, $\mathcal{F}_{\theta_e}(\cdot): \mathbb{R}^{H \times W \times 3} \rightarrow \mathbb{R}^d$ which is then used accordingly in either branch.
\\
\noindent\keypoint{Primary branch}: In addition to the backbone feature extractor, this branch lowers the feature dimension of extracted feature to $d_p$ using a linear layer, $\mathcal{H}_{\theta_p}(\cdot): \mathbb{R}^{d} \rightarrow \mathbb{R}^{d_p}$ for better learning. Geared towards instilling a discriminative knowledge, the model is trained on a cross-modal triplet objective following Eq. \ref{equ:baseTrip}, as:
\begin{equation}
    \begin{aligned}
    \mathcal{L}_\text{Tri}^{\theta_e, \theta_p} =  \max \{0, m + \delta(f_S^{d_p}, f_{P^+}^{d_p}) - \delta(f_S^{d_p}, f_{P^-}^{d_p})\}
    \end{aligned}
    \label{equ:finalTrip}
    \vspace{-0.5cm}
\end{equation}
\\
\noindent\keypoint{Auxiliary branch}: 
Owing to the supervision-less test-time training paradigm, we needed to choose an auxiliary task which (a) is self-supervised, so that it can be performed free of label-cost during inference and (b) can complement the primary task in a way such that the extra features learned provide broader interpretation of input data \cite{liu2019self}. We thus opt for a self-modal reconstruction task for either modality. In both cases, the latent feature is first reduced to a lower $d_a$ dimension.
For sketch, as vector coordinates are available, we perform sketch raster-to-vector decoding. 
\\
\keypoint{Sketch Vectorization:} In vector-format one can use five-element vector $v_t = (x_t, y_t, q^1_t, q^2_t, q^3_t ) \in \mathbb{R}^{T\times 5}$ to represent pen states for stroke-level modelling; $T$ being the sequence length. Essentially, ($x_t, y_t$) denotes the absolute coordinate value in a normalised $H \times W$ canvas, while the last three represent binary one-hot vectors \cite{ha2017neural} of three respective pen-states: pen touching the paper, pen lifted, and end of drawing. Starting with a $d_a$-dimensional sketch feature ($f_S^{d_a}$), a linear-embedding layer obtains the initial hidden state ($h_t |_{t=0}$) of the decoder RNN (${\theta_a^S}$) as: $h_0 = W_h f_S^{d_a}+ b_h$. It is then updated as: ${h_t = \text{RNN}(h_{t-1} ; [f_S^{d_a}, \psi_{t-1}] )}$, where $\psi_{t-1}$ is the last predicted point and $[\cdot]$ signifies concatenation. A fully-connected layer then predicts five-element vectors at every time step: $\psi_t = W_yh_t+ b_y$, where $\psi_t = (x_t, y_t, q^1_t, q^2_t, q^3_t) \in \mathbb{R}^{2+3}$ -- first two logits for coordinates, last three for pen-states. Using $(\hat{x}_t, \hat{y}_t, \hat{q}^1_t, \hat{q}^2_t, \hat{q}^3_t)$ as ground-truth at $t^{th}$ step, mean-square error~\cite{sain2021stylemeup} $\mathcal{L}^{(t)}_\text{(MSE)}= \left \| \hat{x}_t - {x}_t \right \|_2 +  \left \| \hat{y}_t - {y}_t\right \|_2$, and categorical cross-entropy losses~\cite{bhunia2021more} $\mathcal{L}^{(t)}_\text{CE} = -\sum_{i=1}^{3}\hat{q}^i_t \log \Big(\frac{\exp({q^i_t})}{\sum_{j=1}^{3}\exp({q^j_t})}\Big)$ are used to train the absolute coordinate and pen state prediction (softmax normalised) respectively, as:
\vspace{-0.35 cm}
\begin{equation}
\begin{aligned}
\mathcal{L}_\text{rec}^S (\psi_t;\; \theta_e, \theta_a^S) =  &\frac{1}{T}\sum_{t=1}^{T} \Big( \mathcal{L}^{(t)}_\text{MSE} + \mathcal{L}^{(t)}_\text{CE} \Big) 
\end{aligned}
\vspace{-0.1cm}
\label{equ:vec}
\end{equation}

\keypoint{Photo-to-Edgemap Translation}: Edgemap holding a lower domain gap with a sketch (both contain only structural information) than a photo, enables this task to align gradients in favour of a better sketch representation, thus augmenting primary objective better than direct photo-to-photo translation.
An edgemap corresponding to the matching photo is created as $E$ = \texttt{edge}($P^+$) $\in \mathbb{R}^{H\times W\times 3}$ where \texttt{edge}($\cdot$) is a function that extracts an edgemap from a photo using 2D filters on the grey-scaled input image. Our latent positive-photo feature ($f_{P^+}^{d_a}$) is fed to a convolutional decoder $\text{Dec}_{\theta_a^P}(\cdot) : \mathbb{R}^{d_p} \rightarrow \mathbb{R}^{H\times W\times 3}$ to obtain an edgemap $\widehat{E} = \text{Dec}(f_{P^+}^{d_a})$. We thus have our reconstruction loss as : 
\vspace{-0.1cm}
\begin{equation}
    \label{equ:P2E}
    \begin{aligned}
        \mathcal{L}_{\text{rec}}^P (\theta_e, \theta_a^P) = \|\widehat{E} - E\|_2
    \end{aligned}
    \vspace{-0.0cm}
\end{equation}
For notational brevity, at times we use $\theta_a$ = $\{\theta_a^S, \theta_a^P\}$.

\vspace{-0.0cm}
\subsection{Meta- Learning Auxiliary Reconstruction}\label{sec:Meta-Op}
\vspace{-0.05cm} 
\noindent \textbf{Overview:}
Decreasing the test-train data distribution gap especially for sketches which hold unconstrained diversity \cite{sain2021stylemeup} is quite non-trivial a task. Aiming to alleviate it using test-time training alone would be an ambitious goal, if not insufficient. 
% It is thus important to condition the encoder in a similar setup during training itself. 
We thus take to the meta-learning training paradigm \cite{tim2020metaSurvey} where, the goal is to learn good initialization parameters representing an across-task shared knowledge among related tasks, such that it can quickly adapt to any novel task, with a few gradient update iterations. This simulates a test-time training paradigm in the training itself, which thus conditions the encoder to adapt better during inference. We modify a popular optimization-based meta-learning algorithm is model-agnostic meta-learning (MAML) \cite{finn2017MAML} to suit our purpose. 

\vspace{0.1cm}
\noindent \keypoint{Task Sampling:}
In a meta-learning framework \cite{tim2020metaSurvey}, a model is trained from various related labelled tasks. To sample a task $\mathcal{T}_{i} \sim p(\mathcal{T})$ here, we first select a random category $C_i$ out of $M$ categories. Out of all sketch-photo pairs in $C_i$, $N_i$ and $r_i$ pairs are randomly chosen for  meta-training ($\mathcal{D}^{trn}_i$) and meta-validation($\mathcal{D}^{val}_i$) respectively. Training here consists of two nested loops. The inner loop update is performed over $\mathcal{D}^{trn}$ with an aim to minimise the loss in the outer loop over $\mathcal{D}^{val}$. Within every set, hard negatives are chosen from rest $M-1$ categories ensuring completely dissimilar instances.

\vspace{0.1cm}
\noindent \keypoint{Meta-learning stroke-weights:}
Furthermore, as sketch raster-to-vector decoding is a sequential problem, $\mathcal{L}_\text{rec}^S$ involves a summation operation (Eq. \ref{equ:vec}) over the stroke sequence, thus treating every stroke-specific loss equally. Arguably, this task-specific adaptation for sequential reconstruction could be boosted if weight values for each stroke-specific loss are learned, such that the model adapts better with respect to those strokes holding higher semantic significance. Intuitively, our model thus learns an across-task knowledge where given a sketch, properties of certain strokes could be closer to the encoded knowledge of MAML's initialisation parameter \cite{baik2020learning}, to enhance easier retrieval once reconstructed. On the contrary, considerable anomalies could exist among certain strokes that are redundant or distracting during retrieval using average knowledge encapsulated inside MAML's initialisation parameter. Consequently, during outer-loop adaptation, the model update should prioritise optimising with respect to those particular strokes whose semantic importance is more inclined towards the unknown regarding the model's initialization.
% Recent studies have shown meta-learning to provide the flexibility in learning hyper-parameters \cite{finn2017MAML}, parameterized loss functions \cite{chebotar2019meta}, learning rates \cite{li2017meta}, or weight attenuation \cite{baik2020learning} in the meta-learning process itself. 
We thus intend to learn stroke-specific weights for stroke-wise reconstruction loss instead of averaging over all strokes.

\vspace{0.1cm}
\noindent\keypoint{Meta-Optimisation:}
Regarding factors influencing such weights, literature shows that gradients used for adaptation in inner loop hold knowledge \cite{baik2020learning} related to disagreement (\textit{i.e.} this information needs further learning or assimilation during adaptation) against model's initialization parameters. 
Computing gradients for all model-parameters, being quite cumbersome, we calculate gradient of $t^{th}$ stroke-specific reconstruction loss with respect to final decoding step (parameter $\phi$) as $\nabla_{\phi} \mathcal{L}^{S\;(t)}_\text{rec}(\theta_e, \theta_a^S)$. It is then concatenated with gradients of triplet loss (Eq. \ref{equ:mainTTT}) which deals with the full sketch representation with respect to $\phi$ (both gradient matrices being flattened) as $\mathcal{J}_t = \texttt{concat}\big(\nabla_{\phi} \mathcal{L}^{S\;(t)}_\text{rec}(\theta_e, \theta_a^S), \nabla_{\phi} \mathcal{L}_\text{Tri}(\theta_e, \theta_p)\big)$. We posit that gradient of the triplet objective and stroke-specific reconstruction losses guides towards determining how to weigh different stroke-specific losses. We thus pass this $\mathcal{J}_t$ via a network $g_\eta$ predicting a scalar weight value for $t$-th stroke-specific loss as $\eta_t$ = $g_\eta(\mathcal{J}_t)$.
\noindent Here, $g_\eta$ is designed as a 3-layer MLP network having parameters $\eta$, followed by a sigmoid to generate weights.
Eq. \ref{equ:vec} thus becomes:
\vspace{-0.1cm} 
\begin{equation}
\begin{aligned}
\mathcal{L}_\text{rec}^S (\psi_t;\; \theta_e, \theta_a^S) = \frac{1}{T}\sum_{t=1}^{T}  \eta_t \cdot \Big( \mathcal{L}^{(t)}_{MSE} + \mathcal{L}^{(t)}_{CE} \Big) 
\end{aligned}
\label{equ:metaVec} 
\vspace{-0.1cm}
\end{equation}
\noindent{Summing up, we have our inner loop loss and update as,}
\vspace{-0.0cm} 
\begin{equation}\label{equ:inner-update} 
    \begin{aligned}
        \mathcal{L}_\text{trn} (\theta_e, \theta_p, \theta_a) &=  \lambda_\text{Tri} \mathcal{L}_\text{Tri} + \lambda_\text{rec} (\mathcal{L}_\text{rec}^S + \mathcal{L}_\text{rec}^P),\\
        (\theta_e',\theta_p') \leftarrow \; &(\theta_e,\theta_p) - \alpha \nabla_{\Theta}\mathcal{L}_{trn}(\Theta; \; \mathcal{D}^{trn} )
    \end{aligned}
\vspace{-0.0cm}
\end{equation}
where, $\Theta = \{\theta_e,\theta_p,\theta_a\}$, $\alpha$ is the learnable inner loop learning rate and $\lambda_\text{Tri}, \lambda_\text{rec}$ are hyper-parameters determined empirically.
With updated model parameters, the primary objective is computed as the loss over validation set ($\mathcal{D}^{val}$) as $\mathcal{L}_{val}=\mathcal{L}_\text{Tri}(\theta_e', \theta_p'; \; \mathcal{D}^{val} )$ which updates all model parameters. As ($\theta_e', \theta_p'$)  depends on $\theta_e', \theta_p'$ and $\theta_a$ via inner-loop update (Eq. \ref{equ:inner-update}), a higher order gradient needs to be calculated for outer loop optimisation with learning rate $\beta$ as: 
\vspace{+0.0cm}
\begin{equation}\label{equ:outer-update}
\begin{aligned}
(\Theta, \eta, \alpha) \leftarrow (\Theta, \eta, \alpha) - \beta \nabla_{\theta_e', \theta_p', \eta,\alpha} \sum^{\mathcal{D}^{val}}_{T_i} \mathcal{L}_{val}({\theta_e'}^{(i)}, {\theta_p'}^{(i)} )
\end{aligned}
\end{equation}
The model updates by averaging gradients over a meta-batchsize of $B$ sampled tasks.  

\subsection{Test-time Training for SBIR}\label{sec:TTTSBIR}
\noindent{
Once trained, it is now important to align the trained model parameters to the test-data distribution before using them to encode test-sketches for retrieval. 
First, before test-time training starts, the model is adapted to the test-set photo distribution using only the photo-to-edgemap auxiliary branch over a few ($\tau_p$) gradient steps to update the feature extractor to $\theta_e^P$. 
The trained feature extractor ($\theta_e$) encodes test-photo $P^\mathcal{T}$ to $f_{P^\mathcal{T}} = \mathcal{F}_{\theta_e} (S^\mathcal{T})$ and uses it to update itself via auxiliary reconstruction task-loss $\mathcal{L}_\text{rec}^P (P^\mathcal{T};\; \theta_e, \theta_a^P)$ (Eq. \ref{equ:P2E}).
\begin{equation}
    \begin{aligned}
    {\theta_e^P} \leftarrow \theta_e - \alpha^\mathcal{T}& \nabla_{\theta_e, \theta_a^P} \mathcal{L}_\text{rec}^P (\mathcal{D}^{val}_P)
    \end{aligned}
    \label{equ:finalTTT_P}
\end{equation}
This aligns the model parameters to the test-set photo distribution for retrieval. Please note that this step is optional and that one can directly use $\theta_e$ instead, before starting test-time training. Now the photo-updated trained feature extractor ($\theta_e^P$) encodes a test-set query-sketch ($S^\mathcal{T}$), to $f_{S^\mathcal{T}} = \mathcal{F}_{\theta_e^P} (S^\mathcal{T})$. 
The auxiliary sketch-vectoriser obtains reconstruction loss $\mathcal{L}_\mathcal{T}^S (\psi_t^\mathcal{T};\; \theta_e^P, \theta_a^S)$ (Eq. \ref{equ:vec}),
where $\psi_t^\mathcal{T}$ is the vector-representation of $S^\mathcal{T}$. $\mathcal{L}_\mathcal{T}^S$ updates the feature extractor over $\tau_s$ steps, using which corresponding test-sketch feature ($f^\mathcal{T}_S$) is extracted for retrieval, 
\begin{equation}
    \begin{aligned}
    \widehat{\theta_e^P} \leftarrow \theta_e^P - \alpha^\mathcal{T}& \nabla_{\theta_e, \theta_a} \mathcal{L}_\text{T}^S (\theta_e^P, \theta_a, \mathcal{D}^{val})
    \\
    f^\mathcal{T}_S &= \mathcal{F}_{\widehat{\theta_e^P }, \theta_p} (S^\mathcal{T})
    \end{aligned}
    \label{equ:finalTTT}
\end{equation}
where, $\alpha^\mathcal{T}$ is the learning rate. Once evaluated, the feature-extractor is re-initialised with the photo-adapted model parameters ($\theta_e^P$), or directly $\theta_e$ if choosing to skip photo-adaptation, for the next test-sample.
}
 
\vspace{-0.3cm}
\section{Experiments}\label{sec:experiments}
\vspace{-0.1cm}
\noindent \textbf{Datasets:}  For category-level SBIR, we use: (i) Sketchy \cite{sangkloy2016sketchy} (extended) -- contains $75k$ sketches across $125$ categories with about $73k$ images \cite{liu2017deep} in total. Following \cite{yelamarthi2018zero} we split it as 21 testing classes disjoint from rest 104 training classes which are separated as $73:31$ for meta-train : meta-test to avoid photo overlap between Sketchy \cite{sangkloy2016sketchy} and ImageNet \cite{deng2009imagenet} datasets. 
% Furthermore, Sketchy is used for Fine-grained retrieval as well, owing to its fine-grained associations with at least 5 sketches per photo. Similar class-splits are used, and accuracy is reported on top-q retrieval metric.
(ii) TU-Berlin Extension \cite{eitz2012humans} -- contains 250 object categories with 80 free-hand sketches per category. Photo part is extended using 204,489 natural images of the same categories from \cite{zhang2016sketchnet}. Following \cite{dey2019doodle} we keep 30 random classes for testing, while 220 training classes are split randomly as 150 for meta-train and 70 for meta-test.
{Category-level SBIR is evaluated similar to \cite{liu2017deep} using mean average precision (mAP@all) and precision considering top 200 (P@200) retrievals.}

\noindent \textbf{Implementation Details:} A VGG-16 network pre-trained on ImageNet is used as the shared feature extractor with final output dimension $d$ = $512$. The primary branch linear layer projects it to $d_p$ = $64$ for triplet objectives. For auxiliary branch, the photo branch reduces to $d_a^P$ = $128$ before feeding to a decoder consisting of a series of stride-2 convolutions, with \texttt{BatchNormRelu} activation on every convolutional layer except the output that has \texttt{tanh} for activation. For sketch-decoding a GRU decoder of hidden state size 128 is used. Furthermore, we use Adam optimiser in both inner and outer loops with learning rates $\alpha = 0.0005$ (initial) and $\beta=0.0001$ respectively during meta-learning with single-step gradient update. During test-time adaptation learning-rate is empirically set at $0.0001$ for both photo and sketch, with $\tau_s$ = $\tau_p$ = $4$ gradient steps. Hyper-parameters  $\lambda_\text{Tri}, \lambda_\text{rec}$ are empirically set to 0.7 and 0.3 respectively. We use a meta-batch size of 32 and set margin $m$ to 0.3.

\vspace{-.2cm}
\subsection{Competitors}\label{Competitors}
\vspace{-.10cm}
We design several baselines aligned to our motivation from different perspectives to evaluate our framework. 
(i) State-of-the-art ZS SBIR methods (\textbf{SOTA}): 
\textit{ZS-Cross} \cite{yelamarthi2018zero} aligns cross-modal sketch-photo features jointly to generalise onto unseen classes, approximating photo features from given sketches via image-to-image translation.
While \textit{ZS-CCGAN} \cite{dutta2019semantically} uses semantic  representation (word2vec) of class labels to learn a joint manifold capable of semantic transfer to unseen categories in an adversarial paradigm, 
\textit{ZS-GRL} \cite{dey2019doodle} combines similar semantic information of class labels with visual sketch information and trains over a gradient-reversal layer to reduce sketch-photo domain gap.
\textit{ZS-SAKE} \cite{liu2019semantic} employs knowledge-distillation paradigm using teacher signal from an ImageNet pre-trained CNN model and constrained by semantic information from category-labels to retrieve in a Zero-shot setting.
(ii) Test-time training baselines (TTT): Following \cite{sun2020test} we design a baseline following our pipeline, \textit{TTT-Rotation}  with a triplet-loss primary objective and the auxiliary task of rotation angle classification on both sketch-image and photos, without meta-learning. Similarly \textit{TTT-Affine} follows \cite{mummadi2021test} in using affine transformations on input images as auxiliary task for Tes-time-adaptation.
(iii) Meta-Learning Baselines (Meta): \textit{Meta-SN-ZS} simply employs vanilla MAML \cite{finn2017MAML} on top of a simple Siamese-network following \cite{yu2016sketch}, trained via triplet loss in both inner and outer loops, in a zero-shot retrieval framework. It adapts using inner loop updates across retrieval tasks over categories in SBIR and over instances in FG-SBIR frameworks. \textit{Meta-Aux-ZS} is identical to \textit{Meta-SN-ZS} except that it adapts using both the auxiliary task of self-modal image reconstruction (for both sketch and photo branch) and triplet objective to minimise only triplet loss in the outer-loop. No test-time training is involved in either one.

\vspace{-0.25cm}
\begin{table}[!hbt]
\centering
\setlength{\tabcolsep}{3.5pt}
\caption{\normalsize{Comparative results of our model against other methods on Categorical SBIR}}
\label{tab:quantitative_fgsbir}
\vspace{-0.25cm}
\footnotesize
\begin{tabular}{clcccc}
\hline
\multicolumn{2}{c}{\multirow{2}{*}{Methods}} & \multicolumn{2}{c}{Sketchy (ext)} & \multicolumn{2}{c}{TU Berlin (ext)} 
\\ \cline{3-6} 
& & {mAP@all} & {P@200}   & mAP@all & P@200
\\ \hline  
\multirow{4}{*}{SOTA}
& ZS-Cross~\cite{yelamarthi2018zero}    & 0.196 & 0.260 & 0.005 & 0.003 \\
& ZS-CCGAN~\cite{dutta2019semantically} & 0.312 & 0.463 & 0.297 & 0.435 \\
& ZS-GRL~\cite{dey2019doodle}           & 0.334 & 0.358  &0.109 &0.121 \\
& ZS-SAKE~\cite{liu2019semantic}        & 0.526 & 0.598  &0.475 &0.609 \\
\hline
\multirow{2}{*}{B-TTT}
& TTT-Rotation \cite{sun2020test}    & 0.428 & 0.514  &0.337 &0.421 \\
& TTT-Affine \cite{mummadi2021test}  & 0.432 & 0.522  &0.351 &0.456 \\
\hline
\rule{0pt}{2ex} 
\multirow{2}{*}{B-Meta}
& Meta-SN-ZS            & 0.368 & 0.452  &0.276 &0.402 \\
& Meta-Aux-ZS           & 0.401 & 0.475  &0.318 &0.447 \\
\hline 
& Proposed           & \bf 0.575  & \bf 0.624 & \bf 0.507 & \bf 0.648 \\
\hline
\end{tabular}
\vspace{-0.4cm}
\end{table}

\vspace{-0.3cm}
\subsection{Result Analysis and Discussion}\label{analysis}
\vspace{-0.15cm} 
Table \ref{tab:quantitative_fgsbir} shows that methods employing Test-time training mostly surpass Zero-shot SBIR methods. Among them, our method consistently outperforms the other state-of-the arts in retrieval accuracy. \textit{ZS-Cross} \cite{yelamarthi2018zero} with its simplistic cross-modal training paradigm is quickly surpassed over by \textit{ZS-CCGAN} \cite{dutta2019semantically} (by 0.116 mAP@all on Sketchy), as the latter is aided with a cycle consistency loss in an adversarial training paradigm in addition to the guidance from word2vec embeddings of categories -- providing much better generalisability for the \textit{unseen} classes. Although superior, it fails to outperform \textit{ZS-GRL}~\cite{dey2019doodle} due to the latter's usage of the gradient-reversal layer that specifically aims to create a domain-agnostic embedding in addition to semantic class labels towards improving accuracy. However in all these methods catastrophic forgetting is a major issue which unavoidably impacts their performance. \textit{ZS-SAKE} \cite{liu2019semantic} specifically focuses on knowledge preservation to reduce this effect, with the help of a knowledge-distillation paradigm, that aims to preserves the knowledge from pre-trained ImageNet \cite{deng2009imagenet} weights while training on the new dataset. The superior result (0.178 mAP@all more than \textit{ZS-GRL}) demonstrates that original domain knowledge preserved by \textit{ZS-SAKE} is not only maintaining its ability to be adapted back to the original domain but also helping the model to be more generalizable to the unseen target domain.

Coming to the test-time adaptation paradigm, we report the result of two state-of-the-art paradigms naively implemented towards our retrieval objective on the two datasets. \textit{TTT-Rotation} \cite{sun2020test} performs rotation-angle classification as an auxiliary task with the primary task being cross-modal triplet loss objective to adjust to test-data distribution during inference. The problem of catastrophic forgetting is alleviated to an extent due to the shift in focus from learning a domain-invariant mapping to evolving the latent space to adjust the test-distribution. Naturally we see a relative rise in accuracy of 0.94 mAP@all against \textit{ZS-GRL} . \textit{TTT-Affine} \cite{mummadi2021test} having learnable affine transform, enables itself to align the trained parameters towards the test-distribution to a greater extent than TTT-Rotation, thus faring slightly better (0.004) than that in accuracy.
Introducing meta-learning in a zero-shot paradigm on top of basic Siamese network trained on Triplet loss (Meta-SN-ZS) improves existing results over the cross-modal ZS experiment (ZS-Cross) by 0.172 mAP@all on Sketchy \cite{sangkloy2016sketchy} . This is because meta-learning conditions the model to retain and use knowledge acquired across a set of relative tasks to adapt and generalise onto new tasks in a simulated testing scenario. Attaching an auxiliary-task branch to the primary and training it in the inner loop with the primary objective, further improves result (by 0.035) in the Zero-Shot setting proving its potential in this area. Our method combines the best of these worlds to use auxiliary reconstruction task, in a meta-learning training paradigm, aided with test-time adaptation for optimal accuracy. Additionally it meta-learns the stroke-specific weights for reconstruction, towards better enhancing the primary discriminative objective, thus outperforming the existing methods.

\begin{figure*}[!hbt] 
    \centering
        \includegraphics[width=\linewidth]{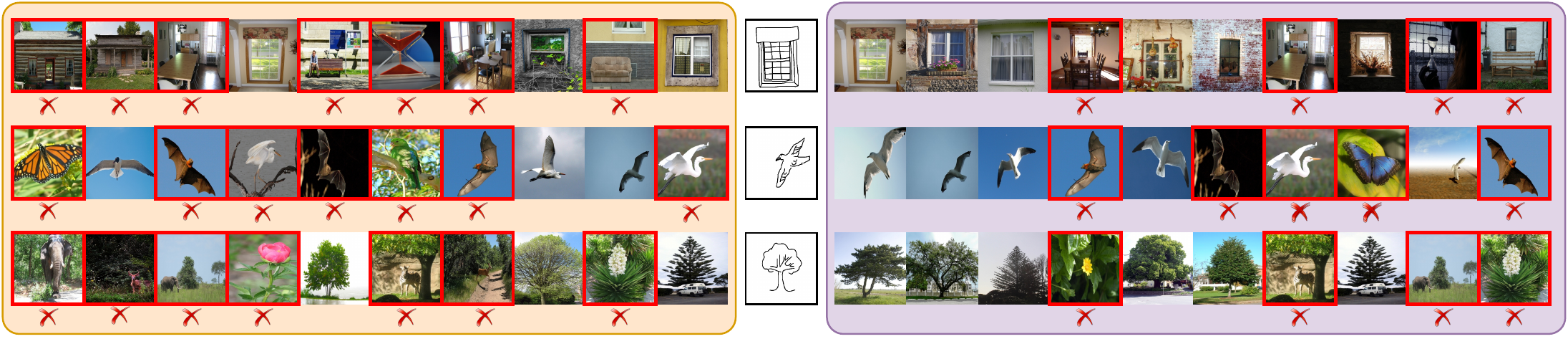}
        \vspace{-0.6cm}
        \caption{Qualitative Zero Shot retrieval results on Sketchy dataset. ZS-Cross (\orange{left}) vs Ours (\violet{right}). }
        \vspace{-0.5cm}
    \label{fig: mainRetrieval}
\end{figure*}
\vspace{-0.2cm}

\subsection{Ablation Study}\label{ablation}
\vspace{-0.18cm}
We perform a detailed ablative study different architectural choice fro various perspectives in Table \ref{tab:abla}.

% \vspace{0.1cm}
\keypoint{\emph{[ii] Is meta-learning important:}}
To judge its contribution we design an experiment training without the meta-learning paradigm in the ZS - setup. The model is trained using two losses (primary and auxiliary) and the auxiliary task updates the model during test-time training. Results (Type-II in Table \ref{tab:abla}) show a stark decrease (by 0.088 mAP@all) against the proposed method, showing how firmly it maintains the discriminative knowledge while training itself, that is otherwise distorted during test-time training. Furthermore using meta-learning avails the option of meta-learning stroke-weights which contributes further.

\keypoint{\emph{[i] Significance of learnable $\mathbf{\eta_t}$ :}}
To show the efficacy of the learnable stroke-specific weight for reconstruction loss, we remove $g_\eta$ simplifying the  sketch-reconstruction loss (Eq. \ref{equ:metaVec}) to MSE and cross-entropy loss (Eq. \ref{equ:vec}). Doing so (Type-III) results in decrease from the proposed method signifying learning relative stroke-importance towards reconstruction is beneficial. Furthermore verifying the dependency of $g_\eta$ on gradients from primary objective, we train a model with $g_\eta$ initialised with a random tensor of fitting dimensions. Without the guidance of supporting the primary objective (discriminative learning), the weights are learned sub-optimally leading to a slight drop by 0.014 mAP@all.

% \vspace{0.1cm}
\keypoint{\emph{[iii] Choice of auxiliary task:}}
One of the most significant aspect of this test-time training paradigm is choosing the auxiliary task -- not only should it be free of label-cost but it must be well suited to capture the test-time distribution over a few gradient updates so as to align model parameters to the test dataset. Without it the model performs quite poorly (Type-I). Exploring other alternatives we thus design a few experiments (Type V-VII) results of which are shown in Table \ref{tab:abla}. Type IV (Img2Img) employs image-to-image translation, decoding the encoded feature via a stride-2 convolutional decoder with \texttt{BatchNormRelu} activation as the auxiliary task, for both branches, i.e sketch-\textit{image} to sketch-\textit{image} and photo-to-photo (not edgemap like ours). Type V performs image-to-image translation on the photo-branch, keeping sketch branch same as ours for the auxiliary task. While Type VI chooses rotation-angle classification as the auxiliary task on both sketch and image branch following \cite{sun2020test}, Type VII employs affine transformation on the photo and sketch-image in either branch, following the auxiliary-task approach of \cite{mummadi2021test}, but keeps the rest of the training paradigm like meta-learning identical to ours. 
\vspace{-0.25cm}
\setlength{\tabcolsep}{4pt}
\begin{table}[!hbt]
    \centering
    \footnotesize
    \caption{Ablative studies (accuracy on Sketchy) .}
    \vspace{-0.2cm}
    \begin{tabular}{cccccccc}
        \hline
        Type & Primary & Auxiliary & Meta & TTT & $\eta$ & mAP@all & P@200\\
        \hline
        I   &  \checkmark & \xmark & \checkmark & - & - & 0.368 & 0.452 \\
        II  &  \checkmark & \checkmark & \xmark & \checkmark & - & 0.487 & 0.576  \\
        % III &  \checkmark & \checkmark & \checkmark & \xmark & \checkmark & 99.99 & 99.99  \\
        III  &  \checkmark & \checkmark & \checkmark & \checkmark & \xmark & 0.561 & 0.610  \\
        IV   &  \checkmark & Img2Img & \checkmark & \checkmark & \checkmark & 0.528 & 0.601  \\
        V  &  \checkmark & Photo-Vec & \checkmark & \checkmark & \checkmark & 0.546 & 0.605  \\
        VI &  \checkmark & Rotation & \checkmark & \checkmark & \checkmark & 0.511 & 0.596  \\
        VII &  \checkmark & Affine & \checkmark & \checkmark & \checkmark & 0.524 & 0.597  \\
        VIII &  \checkmark & Edge-LSTM & \checkmark & \checkmark & \checkmark & 0.568 & 0.619  \\
        IX &  \checkmark & Edge-TF & \checkmark & \checkmark & \checkmark & 0.562 & 0.615  \\
        X &  \checkmark & Edge-Offset & \checkmark & \checkmark & \checkmark & 0.570 & 0.622  \\
        Ours &  \checkmark & \checkmark & \checkmark & \checkmark & \checkmark & 0.575 & 0.624  \\
        \hline
    \end{tabular}
    \label{tab:abla}
    \vspace{-0.2cm}
\end{table}
In context of SBIR, we observe that sketch raster-to-vector translation holds significance as Type V performs better than Type IV. Furthermore, our method's superiority over Type V confirms photo-to-edgemap translation to be a better suitable auxiliary task in context of sketches. While types VI and VII both morph the photo and sketch as images, apparently the classification objective alone isn't sufficiently strong as reconstruction to align model parameters adequately to the test distribution. We also compare efficiency of sketch in terms of vector format -- absolute coordinates (ours) vs. offset-coordinate (Type X) \cite{ha2017neural}. Turns out the former is better for decoding. Comparing sketch decoders between GRU (ours), LSTM (Type VIII) and Transformer (TYPE IX), showed GRU as optimum empirically.

\begin{figure}[!hbt]
\begin{center}
  \includegraphics[width=\linewidth]{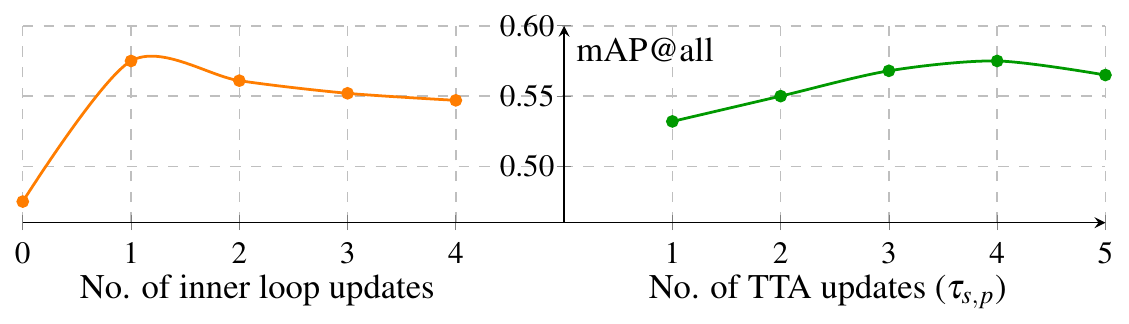}
\end{center}
\vspace{-.65cm}
\caption{Model performs optimally at 1 meta-training gradient update (left) and 4 test-time adaptation updates (right)}.
\vspace{-0.3cm}
\label{fig:steps}
\end{figure}

\begin{figure}[!hbt]
\begin{center}
  \includegraphics[width=\linewidth]{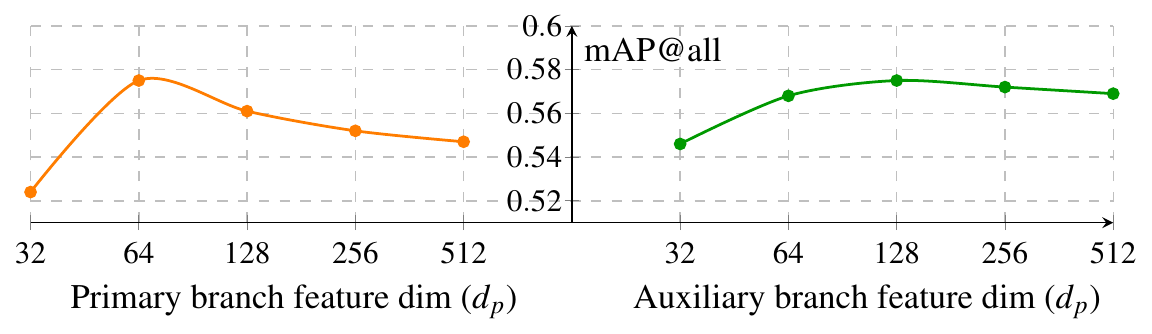}
\end{center}
\vspace{-.6cm}
  \caption{Varying feature dimension for primary objective (left) -- optimal = 64, and for auxiliary decoding (right) -- optimal = 128.
  \vspace{-.4cm}
}
\label{fig:dim}
\end{figure}

\vspace{0.1cm}
\noindent \textbf{\emph{[iv] Further insights:}} 
Qualitative results on Sketchy \cite{sangkloy2016sketchy} are shown in Fig.~\ref{fig: mainRetrieval}.
Fig. \ref{fig:steps} shows that, during training one single adaptation step is found to be optimal with the highest performance gain. Diminishing results on higher updates contradicting \cite{finn2017MAML}, might be due to detrimental concentration of inner loop on irrelevant sketch details, thus forgetting learned generic prior knowledge. During inference however model parameters find four gradient update steps to be optimal for aligning to the test distribution. More steps induce confusion, leading to a drop in accuracy. Furthermore, an ablative study (Fig. \ref{fig:dim}) showed optimal feature dimension for primary and auxiliary objectives to be 64 and 128 respectively, almost retaining performance with higher ones. Also, evaluating our model without the optional one-time update (\S \ref{sec:TTTSBIR}) on test-set gallery photos, we obtain a slight drop in results to 0.560 mAP@all in Sketchy.
Compared to 8.8 ms of ZS-Cross, ours takes 19 ms more per query, due to the additional test-time training involved.
\vspace{-0.3cm}
\vspace{-0.8cm}

\vspace{-0.2cm}
\section{Conclusion}
\vspace{-.15cm}
\noindent{
In this paper we extended the definition of ZS-SBIR, asking it to extend not just to novel categories, but also to new style of query sketches. We achieve this by proposing a test-time training paradigm that adapts the trained model using just one sketch. Firstly, we show that sketch raster-to-vector translation on query-sketch alone is reliable to bridge the train-test gap as an auxiliary task. Secondly, we propose a novel meta-learning paradigm to ensure test-time updates from this auxiliary task would \textit{not} be adversely affecting the joint embedding that is used to conduct retrieval. Extensive experiments with ablative studies show our method to surpass other state-of-the-arts.
}

{\small
\bibliographystyle{ieee_fullname}
\bibliography{main}
}

\cleardoublepage

\title{\vspace{-1.8cm}\Large{\textbf{Supplementary material for \\ Sketch3T: Test-Time Training for Zero-Shot SBIR}\vspace{-0.5cm}}}
% \author{}
% \lstset{
%     basicstyle=\ttfamily,
%     % language=python,
%     keywordstyle=\color{blue},
%     stringstyle=\color{DarkMagenta},
%     commentstyle=\color{DarkGreen},
%     morecomment=[l]{\%}
% }

\author{
Aneeshan Sain\textsuperscript{1,2}  \hspace{.2cm} 
Ayan Kumar Bhunia\textsuperscript{1} \hspace{.3cm}  
Vaishnav Potlapalli$^{*}$  \hspace{.3cm}  
Pinaki Nath Chowdhury\textsuperscript{1,2}   \\  
Tao Xiang\textsuperscript{1,2}\hspace{.2cm}  
Yi-Zhe Song\textsuperscript{1,2} \\
\textsuperscript{1}SketchX, CVSSP, University of Surrey, United Kingdom.  \\
\textsuperscript{2}iFlyTek-Surrey Joint Research Centre on Artificial Intelligence.\\
{\tt\small \{a.sain, a.bhunia, p.chowdhury,  t.xiang, y.song\}@surrey.ac.uk
% ; pvaishnav2718@gmail.com
} 
\vspace{-1.0cm}
}

\maketitle

\section*{Clarity on computational overhead:}

Delving into the complexity analysis of our method we explore complexity of a relevant method in this context. The Table below compares the complexity of ZS-SAKE \cite{liu2019semantic} with ours. ZS-SAKE is indeed simpler to train, and faster at test-time. The extra cost is however justifiable by (i) the ability to handle style changes in addition to novel categories, (ii) we do not dictate word embedding (as per ZS-SAKE), but just a single sketch, and (iii) we surpass ZS-SAKE \cite{liu2019semantic} by a rather significant {9.31\%} margin (relative mAP@all).

\vspace{+0.1cm}
\setlength{\tabcolsep}{4pt}
\begin{table}[!hbt]
    \centering
    \small
    % \caption{Comparison of computational complexity}
    \vspace{-0.2cm}
    \begin{tabular}{ccc}
        \hline
        Method & Parameters  & Time per Forward Pass \\
        \hline
        ZS-SAKE \cite{liu2019semantic}   &  27.6 mil.  & 25.6 ms \\ % & f3.8B & m7.7B
        Ours  &  33.8 mil. & 110.4 ms \\
        \hline
    \end{tabular}
    \label{tab:extra}
    \vspace{-0.2cm}
\end{table}

\section*{Clarity on auxiliary loss used:} 

Without the auxiliary objective, test-time training is infeasible thus dropping model performance (Table \ref{tab:abla}, Type-I in main paper). Analysing further (Type IV-VII), we found reconstructing stroke-level details optimally conditions the encoder to a \textit{sketch}, as it is penalised on stroke-level semantics, proving its superiority in aiding the primary objective. Furthermore, learning which strokes are significant towards boosting the primary task (via $\eta_t$ in Type III) is advantageous, as some strokes inherently hold more semantic meaning in a sketch than others.

\section*{Clarifying experiments:}

Our work differs from \cite{sun2020test} in our latent space preservation via meta-learning, and in our auxiliary task which is optimally suited to sketches.  
Table below compares the performance of \cite{wang2021tent,liang2020we} adjusted for retrieval, against ours. To clarify, in both Tables \ref{tab:quantitative_fgsbir} and \ref{tab:abla}, our method uses test-set photo reconstruction.
% Besides SOTAs, the baselines shadow their respective methods, leaving out the option of extra test-set photo adaptation. 
In Table \ref{tab:abla}, all methods involving test-time training and auxiliary task have employed test-set photo adaptation (TPA) as well. Without it, accuracy dips slightly by 0.020 mAP@all on average.
Table below shows our method's accuracy in that setting (\textbf{Ours w/o TPA}).

\vspace{-0.1cm}
\begin{table}[!hbt]
% \normalsize
\centering
\setlength{\tabcolsep}{3.5pt}
% \caption{\normalsize{Comparative results of our model against other methods on Categorical SBIR}}
\label{tab:quantitative_fgsbir_extra}
% \vspace{-0.25cm}
\small
\begin{tabular}{lcccc}
% \specialrule{1.pt}{1pt}{1pt}
\hline
\multirow{2}{*}{Methods} & \multicolumn{2}{c}{Sketchy (ext)} & \multicolumn{2}{c}{TU Berlin (ext)} 
\\ \cline{2-5} 
 & {mAP@all} & {P@200}   & mAP@all & P@200
\\ \hline  
% \rule{0pt}{3ex}
B-TENT \cite{wang2021tent}    & 0.483 & 0.574 & 0.405 & 0.521 \\
% \rule{0pt}{2ex}
B-SHOT~\cite{liang2020we} & 0.497 & 0.578 & 0.425 & 0.538 \\
% \rule{0pt}{2ex}
Ours w/o TPA           &  0.561  &  0.620 &  0.495 &  0.642 \\
Ours            & \bf 0.575  & \bf 0.624 & \bf 0.507 & \bf 0.648 \\
\hline
\end{tabular}%
\vspace{-0.4cm}
\end{table}

\section*{Sensitivity of hyper-parameters:} 

The initial estimate for some hyper-parameters like margin value of triplet loss, or initial values of inner and outer learning rates were inspired from related works \cite{stylemeup} and optimised empirically thereafter. We have experimented by changing the ratio $\lambda_{Tri}:\lambda_{rec}$ from $7$:$3$ to $1$:$1$ which dipped performance to 0.510 (0.581) mAP@all (P@200) on Sketchy showing a slight sensitivity on the ratio of learning objectives. We shall include such hyperparameter sensitivity details on acceptance. For other ablation studies on sensitivity of the number of gradient steps, of both test-time training and meta-learning, or on optimal feature dimension for primary and auxiliary tasks, please refer to Fig. \ref{fig:steps} and Fig. \ref{fig:dim} respectively, in the main paper.

\section*{Additional visualisations:} {Following diagram shows sketches reconstructed via the decoder (lower) against input (upper). }

\begin{figure}[!hbt]
    \centering
    \includegraphics[width=\linewidth]{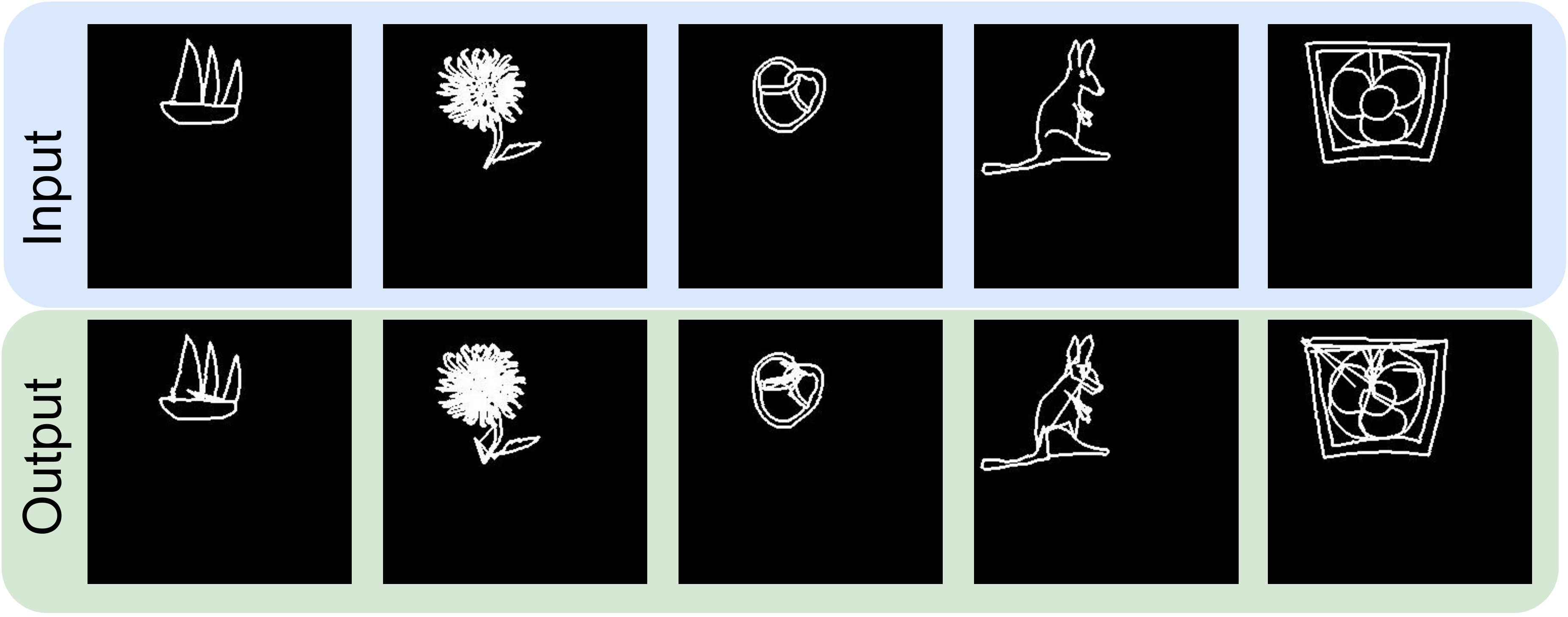}
    % \caption{Caption}
    \label{fig:my_label}
    \vspace{-0.6cm}
    \end{figure}

\section*{Limitations:}  Despite the effective paradigm of our proposed method, there might be some cases, where the model fails to retain its learnt cross-modal knowledge of the source data. As evident from the $4^\text{th}$ sample in Figure above, the sketch reconstructed might indulge certain noisy strokes which infers that the test-time training will not always be optimal for very complex types of sketches.

% \section*{References: \quad}
% \noindent{\textbf{[A]} Do we really need to access the source data? source hypothesis transfer for unsupervised domain adaptation, ICML 2020. \\
% %
% \textbf{[B]} TENT: Fully Test-Time Adaptation by Entropy Minimization, ICLR 2021.}

% {\small
% \bibliographystyle{ieee_fullname}
% \bibliography{Original_egbib}
% }
% \end{document}

\end{document}